\documentclass[sigconf]{acmart}
\usepackage[T1]{fontenc}
\usepackage{threeparttable}
\usepackage{multirow}
\AtBeginDocument{%
	\providecommand\BibTeX{{%
			\normalfont B\kern-0.5em{\scshape i\kern-0.25em b}\kern-0.8em\TeX}}}

\copyrightyear{2021}
\acmYear{2021}
\setcopyright{acmcopyright}\acmConference[ACM MM '21]{Proceedings of the 2021 ACM MULTIMEDIA}{20--24 October 2021, 2021}{Chengdu, China}
\acmBooktitle{Proceedings of the 2021 ACM MULTIMEDIA (ACM MM '21), 20--24 October 2021, Chengdu, China}
\acmPrice{15.00}
\acmDOI{}
\acmISBN{}

\begin{document}
	\fancyhead{}
	\title{Multimodal Feature Fusion for Video Advertisements Tagging Via Stacking Ensemble}
	
\author{Qingsong Zhou}
\email{19s152104@stu.hit.edu.cn}
\affiliation{%
	\institution{Harbin Institute of Technology}
	\city{Shenzhen}
	\postcode{150006}
	\country{China}}

\author{Hai Liang}
\email{lianyhai@gmail.com}
\affiliation{%
	\institution{Hebei university}
	\postcode{071002}
	\country{China}}

\author{Zhimin Lin}
\email{linzhimin.lzm@alibaba-inc.com}
\affiliation{%
	\institution{Alibaba Group}
	\city{Hangzhou, China}
}

\author{Kele Xu}
\email{kelele.xu@Gmail.com}
\authornote{Corresponding author.}
\affiliation{%
	\institution{National University of Defense Technology}
	\city{Changsha}
	\postcode{410073}
	\country{China}}	

\renewcommand{\shortauthors}{Qingsong, et al.}
	
\begin{abstract}
Automated tagging of video advertisements has been a critical yet challenging problem, and it has drawn increasing interests in last years as its applications seem to be evident in many fields. Despite sustainable efforts have been made, the tagging task is still suffered from several challenges, such as, efficiently feature fusion approach is desirable, but under-explored in previous studies. In this paper, we present our approach for Multimodal Video Ads Tagging in the 2021 Tencent Advertising Algorithm Competition. Specifically, we propose a novel multi-modal feature fusion framework, with the goal to combine complementary information from multiple modalities. This framework introduces stacking-based ensembling approach to reduce the influence of varying levels of noise and conflicts between different modalities. Thus, our framework can boost the performance of the tagging task, compared to previous methods. To empirically investigate the effectiveness and robustness of the proposed framework, we conduct extensive experiments on the challenge datasets. The obtained results suggest that our framework can significantly outperform related approaches and our method ranks as the $1st$ place on the final leaderboard, with a Global Average Precision (GAP) of 82.63\%. To better promote the research in this field, we will release our code in the final version.

\end{abstract}



\maketitle

\section{Introduction}

Last decades have witnessed exponentially increasing of the different platforms (e.g. Facebook, TikTok, Flickr, YouTube, etc.), and huge volumes of video advertisements have been posted on the platforms everyday. Online short videos have become indispensable to peoples’ daily lives \cite{wu2014crowdsourced}. Thus, it is desirable to automatically assign the tags to the user-generated video advertisements, which can be further deployed for video ads recommendation \cite{dey2020recommendation,velusamy2008efficient}, and video content retrieval \cite{rossetto2020interactive,lokovc2018influential,ding2019long}.

As a fundamental task in multimodal machine learning, video tagging has drawn lots of attentions and witnessed tremendous progress since last decades \cite{hadar2017implicit}. The applications of video tagging seem to be evident in many fields. However, video ads tagging is still confronted with several challenges, such as the varying levels of noise and conflicts between different modalities. Thus, the tagging task still falls short of accuracy, and sustainable efforts have been made to mitigate aforementioned issues. As demonstrated in previous studies, the tagging task's performance heavily relied on the fusion of multimodal features, and different fusion strategies have been proposed, including the early-stage fusion or later-stage fusion \cite{baltruvsaitis2018multimodal}. Since the revolution of deep neural network, deep learning-based methods have gradually dominated the video tagging task \cite{gordo2016deep,xu2019multimodal}.

As a special case of video tagging, the video advertisements tagging have also drawn increasing research interests \cite{cheng2020fashion,hidayati2020dress}, due to its vast market. Nevertheless, video ads tagging is confronted with more challenges, compared to the conventional video tagging. Firstly, the backgrounds of the videos are more complex and diverse. Secondly, brightness/viewpoint variation in the practical settings can be observed rather different in the videos. Thirdly, the involved video ads often come from multi heterogeneous modalities, ranges from image, sound, and text information. Different modalities exhibit different characteristics in the practical settings \cite{ji2017cross,hidayati2017learning}, thus may leads to conflicts between different modalities. All of aforementioned issues increase the difficulties of the video ads tagging task.

\begin{figure}[h]
	\centering
	\includegraphics[width=8.5cm]{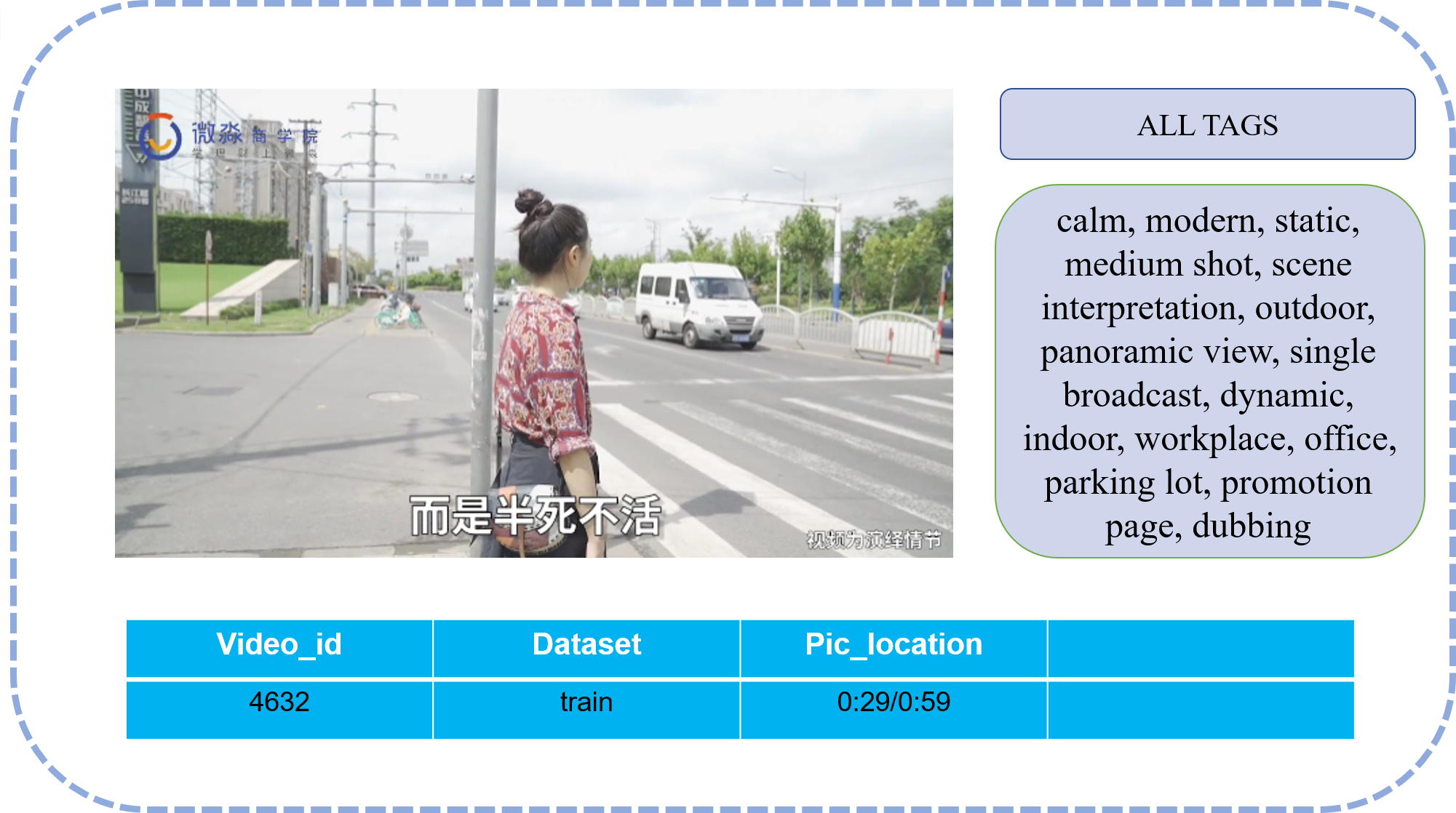}
	\caption{One example of from the challenge dataset. The datasets consists of video id information, video contents, caption and tags.}
	\label{sample}
\end{figure}

In this paper, we explore an efficient multimodal feature fusion framework for the video tagging task, leveraging model stacking \cite{xu2020multimodal,yang2021cross}. The main motivation of our framework is that the knowledge from different modalities can and should be combined, with the goal to enhance the tagging performance. Moreover, the fusion strategy should handle with the conflicts between different modality. With our stacking framework, we can jointly train the tagging model by utilizing the data from multiple modalities. The contributions can be summarized as follows:
1). We propose a novel learning framework to explore knowledge from different modalities for the video tagging task.
2). We comprehensively investigate the effectiveness and robustness of the model for 2021 ACM Multimedia Multi-modal ads video understanding challenge. The results demonstrated the effectiveness and robustness of proposed approaches. Moreover, our approach ranked as the $1_{st}$ place on the final leaderboard.

The remainder of this paper is organized as follows. Related work on video tagging is reviewed in Section 2. Section 3 presents the our methodology while experimental results are shown in Section 4, Section 5 drawn conclusions.

\section{Related Work}

Generally speaking, video tagging can be divided into two main stages: feature extraction from multimodal data and the training of a classification model \cite{xu2020multi}. To improve the tagging performance, sustainable efforts have been made, with the goal to extract powerful features. \cite{siersdorfer2009automatic} proposed an automatic video tagging using content redundancy, while \cite{borth2008keyframe} focused on the key frame extraction from the videos. \cite{yang2013effective} suggested that the tags from the image can be transferred for the video tagging task, and \cite{wang2019knowledge} proposed a multimodal deep regression Bayesian networks for the emotion video tagging. \cite{gelli2015image} proposed to leverage pupillary response to estimate video taggings. \cite{wang2017content} proposed to use the users' multiple physiological responses for video emotion tagging while \cite{sasithradevi2020video} suggested to extract spatio-temporal Radon features from the video for the tagging task. \cite{zhu2019enhanced} proposed to integrate the user behavior and content information for the task.

For the classification model, many of recent literature claim that ensemble learning-based approaches have shown superior performance compared to traditional shallow-architecture-based classification model. For example, \cite{shen2012exemplar} used an random forest model for the tagging of human action pose. To obtain better performance, \cite{leung2011handling} explored the multi-instance learning based approach. \cite{he2019feature} proposed to use gradient boosting trees \cite{hew2020predicts,de2018video} for the task with the goal to reduce computational complexity. Deep neural network-based video tagging is proposed in \cite{yang2016multilayer}. Effective video tagging using multiple modalities remains a challenge despite of the aforementioned progress.

\section{Methodologies}
\subsection{Overview of the Proposed Method}

Our framework for the video tagging task is given in Figure 2. As shown in the figure, the framework consists of two main parts, feature extraction and model part. For a given video ad shared on the platforms, we firstly attempt to extract features from the multimodal datasets, including the visual features, text features, and sound-based features. Based on these multi-modal features, different branches will be used to obtain to generate meta-features. In the final stage, a deep neural network model is trained to predict the video ads, via stacking ensemble. We will explain the components in more details subsequently.

\begin{figure*}[h]
	\centering
	\includegraphics[width=15.4cm]{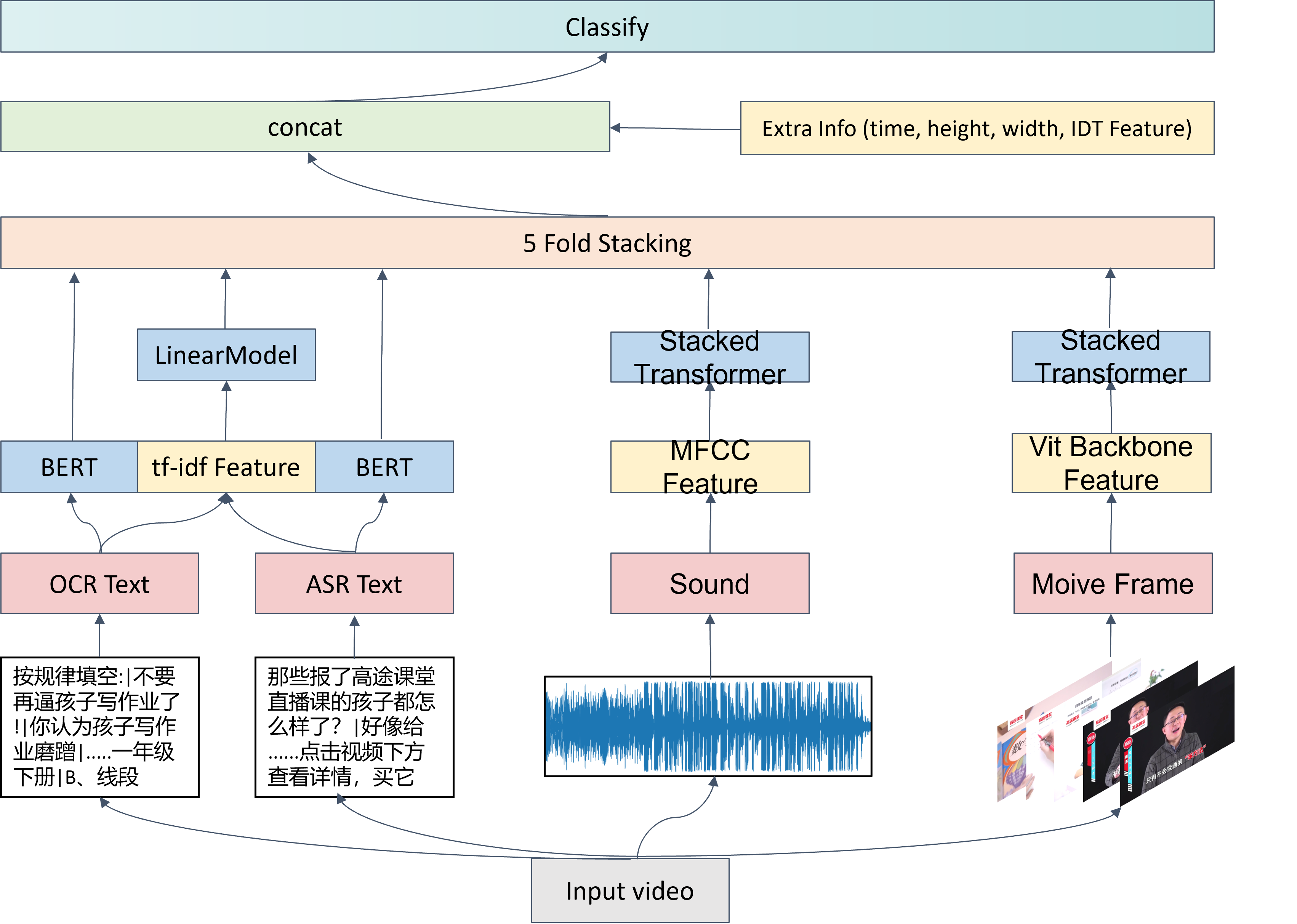}
	\caption{The proposed multimodal stacking framework for the video ads tagging challenge. For the visual feature extraction, we explored the pre-trained vision transformer as our backbone network structure. For the text features, we used the chinese\_bert\_model and tf-idf features, to extract semantic information from the outputs of OCR and ASR. For the sound features, we employ the VGGish architecture for the feature extraction. By combining with extra features (such as, time, height, width of the image), these features are fed into the dense layers together for the tagging task, via stacking ensembling.}
	\label{framework}
\end{figure*}

\subsection{Feature Extraction}

\subsubsection{Text features}

\textbf{Deep text features}. Each video is associated with the text information, including the texts generated from OCR and ASR. 
For the outputs of the ASR, we only selected the first and last 128 words for the training. Instead of using Word2Vec model \cite{church2017word2vec}, we adopt the recent bert\_base\_chinese model to extract the text features \cite{cui2019pre}. Specifically, the BERT model employed a bidirectional training of Transformer, with the goal to learn an efficient language model. BERT has demonstrated its versatile performances on a wide variety of NLP tasks, such as question answering, text classification and machine translation. The term frequency–inverse document frequency (tf-idf) features is also extracted in our framework. Specifically, the jieba word segmentation tool \footnote{https://github.com/fxsjy/jieba} is used in our experiments, and we explore the n-gram(1, 2) after the word segmentation. Finally, we use logistic regression, linearsvc and other linear models for stacking modeling to obtain multi-dimensional modeling information.

\begin{table}[h]
	\centering 
	\caption{Extracted features using the Video Ads Tagging datasets.}
	\begin{tabular}{|c|c|c|}
		\hline
		Type & Pre-trained & Model \\\hline
		tf-idf & No & \\\hline
		OCR features & Yes& BERT \\\hline
		ASR features& Yes& BERT\\\hline
		Sound features & Yes& VGGish\\\hline
		Visual Features& Yes& Transformer\\\hline
	\end{tabular}
	\label{SSC}
\end{table}

\subsubsection{Sound features}

For the sound features, we employ the VGGish architecture for the feature extraction \cite{hershey2017cnn}. The VGGish model was trained on a large-scale audio dataset, thus can be helpful to provide powerful and rich audio representation. For each audio clip, the size of the output is $62 \times 128$.

\subsubsection{Visual features}

Convolutional neural network has become a methodology choice method for the image representations. Recently, the transformer-based models provided impressive results in many vision tasks. We utilize a recent vision transformer (ViT) model \cite{dosovitskiy2020image}, which is pre-trained on ImageNet for the visual features extraction. We resized the image to the size of resized to 384 $\times$ 384. 
We set the number of input layers as 24, while the hidden size is 1024. For the frame extraction from the video, and the image frame features are extracted in 500 ms per frame.

\subsection{Deep Neural Network-Based Model stacking}
By ensembling diverse models can greatly improve the accuracy and robustness for the tagging task. However, the ensemble learning has been under-explored for the multimodal machine learning, due to the heterogeneity of multimodal data. In this paper, we explore the use of stacking of different modalities to improve accuracy and robustness for the video tagging task. Fig. 2 shows the proposed fusion architecture, which is composed of two levels. By random split the data into 5 folds, we build the tagging model for each modality separately. The predicted probabilities of different classes will be concatenated to generate meta-features. Except for the single modal-based meta features, we also employ extra features for the tagging task. For the ensemble learning, we employ the 3-layers deep neural network in our experiments. To prevent overfitting, the dropout strategy is employed in our settings.

\section{Experiments}

\subsection{Dataset and Experimental settings}
We conduct the extensive experiments on the challenge dataset. The performance evaluation are conducted on the provided validation datasets. For the quantitative comparison between different models, the Global Average Precision (GAP) is adopted as performance metrics. For each ad video, our submission consists of a list of predicted labels and the corresponding confidence scores.
To calculate the GAP, the evaluation employs the prediction label with the highest $k(i)$, where $i$ is the index, the confidence score of each video under each index. Suppose that: the video has $k(i)$ predictions (marker/confidence pairs) and can be sorted by its confidence score, then GAP can be formulated as:
\begin{equation}
	GAP = \sum_{i=1}^{3} \sum_{j=1}^{k(i)} p(j)\Delta r(j)
\end{equation}
Larger value of GAP score is preferred, which indicates better tagging performance.

\subsection{Evaluation of Our Method}
We followed the same setting for quantitative comparison, excepted been highlighted in specifics settings. Table 2 presents the results of the tagging task using one single modality. Accuracy and GAP are used to evaluate the performance. As can be seen from the table, visual features can provide higher accuracy and GAP, while sound feature's performance is the worst. These results suggest that visual features may be better utilized the tagging task. We also conduct the experiments leveraging different feature combinations, and the results are given in Table 3. As can be seen from the table, all the combinations can provide a satisfactory result. Moreover, we observe that adding more features can always provide increasing accuracy combined with other features, which demonstrate the importance of feature extraction.

\begin{table}[]
	\caption{Performance comparison between single Modality.}
	\begin{tabular}{|l|l|l|}
		\hline
		Methodology    &  Accuracy &  GAP \\ \hline
		ASR & 0.9169       & 0.74614  \\ \hline
		OCR & 0.9182       & 0.75612  \\ \hline
		Visual   & 0.9257       & 0.79636   \\ \hline
		Sound   & 0.9086       & 0.70701  \\ \hline
	\end{tabular}
\end{table}

To further demonstrate the performance of our framework, we conducted a quantitative comparison with different feature fusion approaches, which includes: (1) Simple feature concat; (2) Sum-pooling for the features; (3) Max-pooling for the features;
(4) Attention mechanism. It is worthwhile to notice, we follow the parameters setting except the fusion settings and the evaluation metrics are presented in Table 4. As can be seen from the table, our feature fusion strategy exhibits better performance. On the other hand, our deep neural network- stacking framework outperforms the other methods proposed, when the accuracy and GAP metrics are used for evaluation. The superior performance also verified the effectiveness of the proposed multi-modal feature extraction module.

\begin{table}[]
	\caption{Performance comparison between different feature combinations.}
	\begin{tabular}{|l|l|l|}
		\hline
		Methodology                             &  Accuracy &  GAP \\ \hline
		Visual+Sound                        & 0.92700       & 0.80018    \\ \hline
		Visual+Text                         & 0.93000         & 0.81569     \\ \hline
		Visual+Sound+Text                     & 0.93100        & 0.81659   \\ \hline
		Visual+Sound+Text +tfidf               & 0.93340       & 0.82641   \\ \hline
		Visual+Sound+Text+tfidf+extra & 0.93350       & 0.82639  \\ \hline
	\end{tabular}
\end{table}

\begin{table}[]
	\caption{Performance comparison between different fusion strategy.}
\begin{tabular}{|l|l|l|}
\hline
Methodology                             &  Accuracy &  GAP \\ \hline
Our method & 0.9310       & 0.81659 \\ \hline
Feature concat                 & 0.91820       & 0.80971 \\ \hline
Sum-pooling                       & 0.9270       & 0.80197 \\ \hline
Max-pooling                    & 0.92750       & 0.80511 \\ \hline
Attention                 & 0.92930       & 0.81295 \\ \hline
\end{tabular}
\end{table}

To further demonstrate the effectiveness our method, we randomly select one video samples from the validation set for visualizations. As shown in Figure 3, the ground-truth and predicted tagging are given using our method. As can observed, the true positive tagging are be found in top-9 results, which demonstrates its effectiveness.

\begin{figure}[h]
	\centering
	\includegraphics[width=8cm]{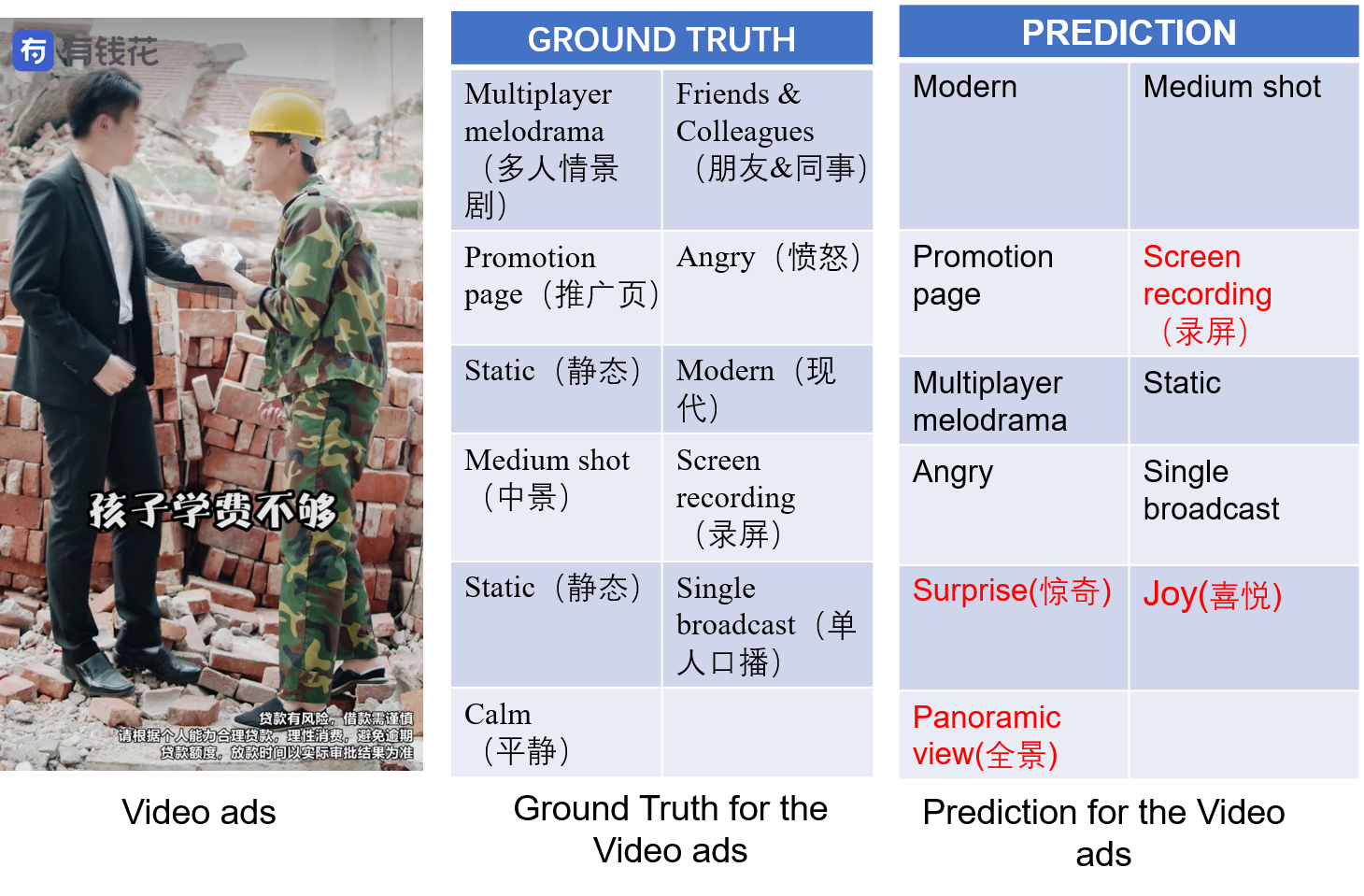}
	\caption{The visualization of predictions, using proposed framework.}
	\label{predicted}
\end{figure}

\section{Conclusion}

In this paper, we propose a novel multimodal feature fusion strategy for the video tagging task. We consider different perspectives (visual, text, sound, and other features) to predict the tagging of posts, and train a DNN-based classification model to obtain the taggings. Extensive experimental results demonstrate that our method can provide better performance, with compared other fusion method. Our team ranked as $1st$ place in the final leaderboard of the video tagging challenge. Both univariate and ablation studies provide useful insights regarding the importance of these features. For our future work, there are several promising directions, such as, leveraging the proposed framework, it would be interesting to fine-tune the entire process using the end-to-end manner.

\section{Acknowledgments}
This work is supported by the major Science and Technology Innovation 2030 "New Generation Artificial Intelligence" project 2020AAA0104803.

\bibliographystyle{ACM-Reference-Format}
\bibliography{sample-base}

\end{document}